\documentclass[11pt,a4paper]{article}
\usepackage[hyperref]{acl2019}
\usepackage{times}
\usepackage{latexsym}
\usepackage{graphicx} 
\usepackage{amsmath}
\usepackage{tikz}
\usepackage{bayesnet}
\usepackage{amssymb}
\usepackage{booktabs}
\usepackage{multirow}
\usepackage[capbesideposition=outside,capbesidesep=quad]{floatrow}

\usepackage{amsmath,amsfonts,bm}

\def\eqref#1{equation~\ref{#1}}

\def\1{\bm{1}}

\def\vmu{{\bm{\mu}}}

\def\vx{{\bm{x}}}
\def\vy{{\bm{y}}}
\def\vz{{\bm{z}}}

\def\mP{{\bm{P}}}

\def\mS{{\bm{S}}}

\def\mU{{\bm{U}}}
\def\mV{{\bm{V}}}
\def\mW{{\bm{W}}}

\def\mSigma{{\bm{\Sigma}}}

\DeclareMathAlphabet{\mathsfit}{\encodingdefault}{\sfdefault}{m}{sl}
\SetMathAlphabet{\mathsfit}{bold}{\encodingdefault}{\sfdefault}{bx}{n}

\DeclareMathOperator*{\argmax}{arg\,max}

\usepackage{url}
\usepackage{array}

\usepackage{selinput}
\SelectInputMappings{%
  aacute={á},
  ntilde={ñ},
  Euro={€}
}

\aclfinalcopy 

\newcolumntype{?}{!{\vrule width 2pt}}

\title{Multilingual Factor Analysis}

\author{
  Francisco Vargas, Kamen Brestnichki, Alex Papadopoulos-Korfiatis \and Nils Hammerla \\ \texttt{Babylon Health} \\
  \texttt{\{firstname.lastname, alex.papadopoulos\}@babylonhealth.com}
}

\date{}

\begin{document}
\maketitle
\begin{abstract}
  In this work we approach the task of learning multilingual word representations in an offline manner by fitting a generative latent variable model to a multilingual dictionary. We model equivalent words in different languages as different views of the same word generated by a common latent variable representing their latent lexical meaning. We explore the task of alignment by querying the fitted model for multilingual embeddings achieving competitive results across a variety of tasks. The proposed model is robust to noise in the embedding space making it a suitable method for distributed representations learned from noisy corpora.
\end{abstract}

\section{Introduction}
Popular approaches for multilingual alignment of word embeddings base themselves on the observation in \cite{mikolov2013exploiting}, which noticed that continuous word embedding spaces \citep{mikolov2013distributed,pennington2014glove, bojanowski2017enriching, JoulinGBM16fasttext} exhibit similar structures across languages. This observation has led to multiple successful methods in which a direct linear mapping between the two spaces is learned through a least squares based objective \cite{mikolov2013exploiting, smith2017offline,xing2015normalized} using a paired bilingual dictionary.

An alternate set of approaches based on Canonical Correlation Analysis (CCA) \cite{knapp1978canonical} seek to project monolingual embeddings into a shared multilingual space \citep{faruqui2014improving, lu2015deep}. Both these methods aim to exploit the correlations between the monolingual vector spaces when projecting into the aligned multilingual space. The multilingual embeddings from \citep{faruqui2014improving,lu2015deep} are shown to improve on word level semantic tasks, which sustains the authors' claim that multilingual information enhances semantic spaces.

In this paper we present a new non-iterative method based on variants of factor analysis \citep{browne1979maximum,mcdonald1970three,browne1980factor} for aligning monolingual representations into a multilingual space. Our generative modelling assumes that a single word  translation pair is generated by an embedding representing the lexical meaning of the underlying concept. We achieve competitive results across a wide range of tasks compared to state-of-the-art methods, and we conjecture that our multilingual latent variable model has sound generative properties that match those of psycholinguistic theories of the bilingual mind \cite{weinreich1953languages}. Furthermore, we show how our model extends to more than two languages within the generative framework which is something that previous alignment models are not naturally suited to, instead resorting to combining bilingual models with a pivot as in \cite{ammar2016massively}.

Additionally the general benefit of the probabilistic setup as discussed in \cite{tipping1999probabilistic} is that it offers the potential to extend the scope of conventional alignment methods to model and exploit linguistic structure more accurately. An example of such a benefit could be modelling how corresponding word translations can be generated by more than just a single latent concept. This assumption can be encoded by a mixture of Factor Analysers \cite{ghahramani1996algorithm} to model word polysemy in a similar fashion to \cite{athiwaratkun2017multimodal}, where mixtures of Gaussians are used to reflect the different meanings of a word.

The main contribution of this work is the application of a well-studied graphical model to a novel domain, outperforming previous approaches on word and sentence-level translation retrieval tasks. We put the model through a battery of tests, showing it aligns embeddings across languages well, while retaining performance on monolingual word-level and sentence-level tasks. Finally, we apply a natural extension of this model to more languages in order to align three languages into a single common space.

\section{Background}

Previous work on the topic of embedding alignment has assumed that alignment is a directed procedure --- i.e. we want to align French to English embeddings. However, another approach would be to align both to a common latent space that is not necessarily the same as either of the original spaces. This motivates applying a well-studied latent variable model to this problem.

\subsection{Factor Analysis}

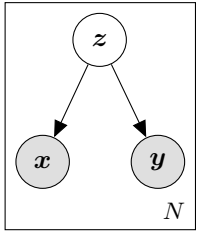
\begin{figure}[t!]
\begin{center}
\begin{tabular}{c}
\hspace{1cm}
\begin{tikzpicture}[x=1.8cm,y=0.9cm]
  \node[obs] (y)  {$\vy$};
  \node[obs, left=of y, xshift=1cm] (x) {$\vx$}; 
  \node[latent, above=of x, xshift=0.75cm] (z) {$\bm{z}$};
  
  \edge {z} {x}; 
  \edge {z} {y}; 

   \plate {} { 
        (x)
        (y)
        (z)
    }{$N$}
\end{tikzpicture}

\end{tabular}
\caption{
  Graphical model for alignment. Latent space $\bm{z}$ represents the aligned shared space between the two vector spaces $\vx$ and $\vy$.} 
 \label{fig:IBFA}
\end{center}
\end{figure}

Factor analysis \citep{spearman1904general, thurstone1931multiple} is a technique originally developed in psychology to study the correlation of latent factors $\vz \in \mathbb{R}^{k}$ on observed measurements $\vx \in \mathbb{R}^{d}$. Formally:
\begin{align*}
    p(\vz) &= \mathcal{N}(\vz ; \bm{0}, \mathbb{I}), \\
    p(\vx | \vz) &= \mathcal{N}(\vx ; \mW \vz + \vmu , \bm{\Psi} ). 
\end{align*}
In order to learn the parameters $\mW ,\bm{\Psi}$ of the model we maximise the marginal likelihood $ p(\vx | \mW, \bm{\Psi})$ with respect to $\mW,\bm{\Psi}$. The maximum likelihood estimates of these procedures can be used to obtain latent representations for a given observation $\mathbb{E}_{p(\vz | \vx)}[\vz]$. Such projections have been found to be generalisations of principal component analysis \cite{pearson1901liii} as studied in \cite{tipping1999probabilistic}.

\subsection{Inter-Battery Factor Analysis}

Inter-Battery Factor Analysis (IBFA) \cite{tucker1958inter, browne1979maximum} is an extension of factor analysis that adapts it to two sets of variables $\vx \in \mathbb{R}^{d}, \vy \in \mathbb{R}^{d'}$ (i.e. embeddings of two languages). In this setting it is assumed that pairs of observations are generated by a shared latent variable $\vz$
\begin{align}
    p(\vz) &= \mathcal{N}(\vz ; \bm{0}, \mathbb{I}), \nonumber \\
    p(\vx | \vz) &= \mathcal{N}(\vx ; \mW_x \vz + \vmu_x  , \bm{\Psi}_x ), \nonumber \\
    p(\vy | \vz) &= \mathcal{N}(\vy ; \mW_y \vz + \vmu_y  , \bm{\Psi}_y ). \label{eq:generative}
\end{align}
As in traditional factor analysis, we seek to estimate the parameters that maximise the marginal likelihood
\begin{align}
    &\argmax_{\{\bm{\Psi}_i, \mW_i\}} \prod_k p(\vx^{(k)}, \vy^{(k)} |  \{\bm{\Psi}_i, \mW_i\}_i),\nonumber   \\
    &\text{subject to  }\bm{\Psi}_i \succ  \bm{0},\: (\mW_i^\top\mW_i)\succcurlyeq  \bm{0} ,\label{objective1}
\end{align}
where the joint marginal $p(\vx_k, \vy_k | \{\bm{\Psi}_i, \mW_i\}_i)$ is a Gaussian with the form
\begin{align*}
    \mathcal{N}&\left(
    \begin{bmatrix}
        \bm { x } \\
        \bm { y }
    \end{bmatrix};
    \begin{bmatrix}
         \bm { \vmu_x }  \\
         \bm { \vmu_y }
    \end{bmatrix},
    \setlength\arraycolsep{0.5pt}
        \begin{bmatrix}
              \mSigma_{xx}  &   \mSigma_{xy}    \\
              \mSigma_{yx}  &  \mSigma_{yy} 
       \end{bmatrix}\!\right), \\ 
      &\:\:\:\:\mSigma_{ij} = \mW_{i}\mW_{j}^{\top} + \delta_{ij} \bm{\Psi}_i,
\end{align*}

\noindent and $\bm{\Psi} \succ \bm{0}$ means $\bm{\Psi}$ is positive definite.

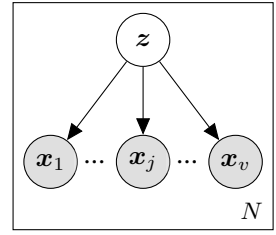
\begin{figure}[t!]
\begin{center}
\begin{tabular}{c}
\hspace{1cm}
\begin{tikzpicture}[x=1.8cm,y=0.9cm]
  \node[obs] (y)  {$\vx_{v}$};
  \node[obs, left= of y, xshift=1.3cm] (x)  {$\vx_{j}$}; 
  \node[obs, left= of x, xshift=1.3cm] (w)  {$\vx_{1}$}; 
  \node[latent, above=of x]    (z)  {$\bm{z}$};
  \node [left= of x, xshift=1cm,fill=white,opacity=.2,text opacity=1] {...};
   \node [left= of y, xshift=1cm,fill=white,opacity=.2,text opacity=1] {...};
  
  \edge {z}   {x}; 
  \edge {z}   {w}; 
  \edge {z}   {y}; 

   \plate {} { 
        (w)
        (x)
        (y)
        (z)
    }{$N$}
\end{tikzpicture}
\end{tabular}
\caption{
  Graphical model for MBFA. Latent space $\bm{z}$ represents the aligned shared space between the multiple vector spaces $\{\vx_{j}\}_{j=1}^{v}$.}.
 \label{fig:MBFA}
\end{center}
\end{figure}
Maximising the likelihood as in Equation \ref{objective1} will find the optimal parameters for the generative process described in Figure \ref{fig:IBFA} where one latent $\vz$ is responsible for generating a pair $\vx, \vy$. This makes it a suitable objective for aligning the vector spaces of $\vx, \: \vy$ in the latent space. In contrast to the discriminative directed methods in \citep{mikolov2013exploiting,smith2017offline,xing2015normalized}, IBFA has the capacity to model noise.

We can re-interpret the logarithm of Equation \ref{objective1} (as shown in Appendix \ref{apdx:reconstruction}) as
\begin{align}
    & \sum_k\!\log p(\vx^{(k)},\!\vy^{(k)} | \bm{\theta})\!=\! C\!+\!\sum_k (\mathcal{L}_k^{y|x}\!+\!\!\mathcal{L}_k^{x}), \label{Mahalanobis} \\
    &\mathcal{L}_k^{y|x} = -\frac{1}{2}|\!|\tilde{\vy}^{(k)} - \mW_{y}\mathbb{E}_{p(\vz|\vx^{(k)})}[\vz] |\!|^2_{\mSigma_{\vy|\vx}}, \nonumber \\
    &\mathcal{L}_k^{x} = -\frac{1}{2}|\!|\tilde{\vx}^{(k)} - \mW_{x}\mathbb{E}_{p(\vz|\vx^{(k)})}[\vz] |\!|^2_{\bm{\Psi_x}\mSigma_{\vx}^{-1}\bm{\Psi_x}}, \nonumber \\
    &C = - \frac{N}{2}(\log |2 \pi \mSigma_{\vy|\vx}| + \log |2 \pi \mSigma_{\vx}|). \nonumber 
\end{align}

The exact expression for $\bm{\Sigma}_{\vy|\vx}$ is given in the same appendix. This interpretation shows that for each pair of points, the objective is to minimise the reconstruction errors of $\vx$ and $\vy$, given a projection into the latent space $\mathbb{E}_{p(\vz|\vx_k)}[\vz]$. By utilising the symmetry of Equation \ref{objective1}, we can show the converse is true as well --- maximising the joint probability also minimises the reconstruction loss given the latent projections $\mathbb{E}_{p(\vz|\vy_k)}[\vz]$. Thus, this forces the latent embeddings of $\vx_k$ and $\vy_k$ to be close in the latent space. This provides intuition as to why embedding into this common latent space is a good alignment procedure. 

In \citep{browne1979maximum,bach2005probabilistic} it is shown that the maximum likelihood estimates for $\{\bm{\Psi}_i, \mW_i\}$ can be attained in closed form
\begin{align*}
    \hat{\mW}_{i} &= \mS_{ii} \mU_{i} \mP^{1/2}, \\
    \hat{\bm{\Psi}}_i &= \mS_{ii} - \hat{\mW}_{i}\hat{\mW}_{i}^{\top}, \\
    \hat{\vmu}_x &= \bar{\vx}, \:\hat{\vmu}_y = \bar{\vy},
\end{align*}
where
\begin{align*}
    \mS_{xx} &= \frac{1}{m}\sum\limits_{i=1}^m \tilde{\bm{x}}^{(i)} \tilde{\bm{x}}^{(i)\top}, \nonumber \\
    \mS_{yy} &= \frac{1}{m}\sum\limits_{i=1}^m \tilde{\bm{y}}^{(i)} \tilde{\bm{y}}^{(i)\top}, \nonumber \\
    \mU_{i} &= \mS_{ii}^{-1/2}\mV_{i},\\ \nonumber
    \mV_{x}\mP\mV_{y}^{\top} &= \text{SVD}(\mS_{xx}^{-1/2} \mS _{xy} \mS_ {yy} ^{-1/2}).
\end{align*}

The projections into the latent space from $\vx$ are given by (as proved in Appendix \ref{apdx:marginal})
\begin{align}
    \mathbb{E}_{p(\vz|\vx)}[\vz] &= (\mathbb{I} + \mW_{x}^{\top}\bm{\Psi}^{-1}_{x}\mW_{x} )^{-1}\bm{W}_{x}^{\top}\bm{\Psi}^{-1}_{x}\tilde{\vx},  \nonumber \\
    \tilde{\vx} &= \vx - \vmu_{x}. \label{eq:moore}
\end{align}

Evaluated at the MLE, \cite{bach2005probabilistic} show that Equation \ref{eq:moore} can be reduced to
\begin{align*}
    \mathbb{E}_{p(\vz | \vx)}[\vz] = \mP^{1/2} \mU_x^{\top} (\vx -\vmu_x ).
\end{align*}

\subsubsection{Multiple-Battery Factor Analysis} \label{subsec:mbfa}

Multiple-Battery Factor Analysis (MBFA) \citep{mcdonald1970three, browne1980factor} is a natural extension of IBFA that models more than two views of observables (i.e. multiple languages), as shown in Figure \ref{fig:MBFA}.

Formally, for a set of views $\{\vx_{1}, ..., \vx_{v}\}$, we can write the model as
\begin{align*}
    p(\vz) &= \mathcal{N}(\vz ; \bm{0}, \mathbb{I}), \\
    p(\vx_{i} | \vz) &= \mathcal{N}(\vx_{i} ; \mW_i \vz + \vmu_i, \bm{\Psi}_i ).
\end{align*}
Similar to IBFA the projections to the latent space are given by Equation \ref{eq:moore}, and the marginal yields a very similar form
\begin{align*}
\setlength\arraycolsep{0.1pt}
    \mathcal{N}\!\left(\!
    \begin{bmatrix}
        \!\vx_{1}\! \\
        \!\vdots\! \\
        \!\vx_{v}\!
    \end{bmatrix}\!;
    \!\begin{bmatrix}
        \!\vmu_1\!  \\
        \!\vdots\! \\
        \!\vmu_v\!
    \end{bmatrix}\!\!,\!\!
    \setlength\arraycolsep{0.1pt}
        \begin{bmatrix}
             \! \mW_{1}\mW_{1}^{\top}\!\!+\!\bm{\Psi}_1  &    \dots\!&\!\mW_{1}\mW_{v}^{\top}   \!\\
              \!\vdots & \ddots  & \vdots \\
              \mW_{v}\mW_{1}^{\top}\:& \hdots & \!\mW_{v}\mW_{v}^{\top}\!\!+\!\bm{\Psi}_v \!
       \end{bmatrix}\!\right)\!\!.
\end{align*}
Unlike IBFA, a closed form solution for maximising the marginal likelihood of MBFA is unknown. Because of this, we have to resort to iterative approaches as in \cite{browne1980factor} such as the natural extension of the EM algorithm proposed by \cite{bach2005probabilistic}. Defining
\begin{align*}
    \bm{M}_t &= \left( \mathbb{I} + \bm{W}_t^\top \bm{\Psi}_t^{-1}\bm{W}_t\right)^{-1}\!\!, \nonumber \\
    \bm{B}_t &= \bm{M}_t\bm{W}_t^{\top}\bm{\Psi}_t^{-1}, \\
    \widetilde{\bm{\Psi}}_{t+1} &=\mS - \mS \bm{\Psi}_t^{-1} \bm{W}_t \bm{M}_t^{\top} \bm{W}_{t+1}^{\top},
\end{align*}
the EM updates are given by
\begin{align*}
    \bm{W}_{t+1}&=\!\!\mS\bm{B}_t^\top \!\!\left( \bm{M}_t + \bm{B}_t\mS\bm{B}_t^\top \right)^{-1}, \\
    \bm{\Psi}_{t+1}\!&=\!\text{Bdiag}\left(\!(\widetilde{\bm{\Psi}}_{t+1})_{11}, \hdots, (\widetilde{\bm{\Psi}}_{t+1})_{vv}\!\right),
\end{align*}
\noindent where $\mS$ is the sample covariance matrix of the concatenated views (derivation provided in Appendix \ref{apdx:em}).
\cite{browne1980factor} shows that, under suitable conditions, the MLE of the parameters of MBFA is uniquely identifiable (up to a rotation that does not affect the method's performance). We observed this in an empirical study --- the solutions we converge to are always a rotation away from each other, irrespective of the parameters' initialisation. This heavily suggests that any optimum is a global optimum and thus we restrict ourselves to only reporting results we observed when fitting from a single initialisation. The chosen initialisation point is provided as Equation (3.25) of \cite{browne1980factor}.
\begin{figure}[t!]
    \centering
    \includegraphics[width=.6\textwidth]{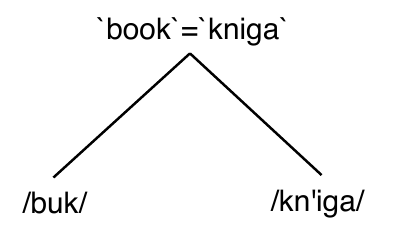}
    \caption{Weinrich's compound model for lexical association between English and Russian. Image from \cite{neuser2017source}.}
    \label{fig:compound}
\end{figure}
\section{Multilingual Factor Analysis}

We coin the term Multilingual Factor Analysis for the application of methods based on IBFA and MBFA to model the generation of multilingual tuples from a shared latent space. We motivate our generative process with the compound model for language association presented by \cite{weinreich1953languages}. In this model a lexical meaning entity (a concept) is responsible for associating the corresponding words in the two different languages.

We note that the structure in Figure \ref{fig:compound} is very similar to our graphical model for IBFA specified in Figure \ref{fig:IBFA}. We can interpret our latent variable as the latent lexical concept responsible for associating (generating) the multilingual language pairs. Most theories that explain the interconnections between languages in the bilingual mind assume that ``while phonological and morphosyntactic forms differ across languages, meanings and/or concepts are largely, if not completely, shared'' \citep{pavlenko2009conceptual}. This shows that our generative modelling is supported by established models of language interconnectedness in the bilingual mind.

Intuitively, our approach can be summarised as transforming monolingual representations by mapping them to a concept space in which lexical meaning across languages is aligned and then performing retrieval, translation and similarity-based tasks in that aligned concept space.

\subsection{Comparison to Direct Methods}

Methods that learn a direct linear transformation from $\vx$ to $\vy$, such as \cite{mikolov2013exploiting, artetxe2016learning, smith2017offline, conneau2017word} could also be interpreted as maximising the conditional likelihood
\begin{align*}
    \prod_k p(\vy^{(k)} | \vx^{(k)}) \!=\!\!\prod_k \mathcal{N}(\vy^{(k)} ;\mW \vx^{(k)}\!+\!\vmu, \bm{\Psi}) .
\end{align*}

As shown in Appendix \ref{apdx:ls}, the maximum likelihood estimate for $\mW$ does not depend on the noise term $\bm{\Psi}$. In addition, even if one were to fit $\bm{\Psi}$, it is not clear how to utilise it to make predictions as the conditional expectation
\begin{align*}
    \mathbb{E}_{p(\vy | \vx^{(k)})}[\vy] = \mW \vx^{(k)} + \bm{\mu},
\end{align*}
does not depend on the noise parameters. As this method is therefore not robust to noise, previous work has used extensive regularisation (i.e. by making $\bm{W}$ orthogonal) to avoid overfitting.

\subsection{Relation to CCA}

CCA is a popular method used for multilingual alignment which is very closely related to IBFA, as detailed in \cite{bach2005probabilistic}. \cite{barber2012bayesian} shows that CCA can be recovered as a limiting case of IBFA with constrained diagonal covariance $\bm{\Psi}_x = \sigma_x^2\mathbb{I}, \: \bm{\Psi}_y = \sigma_y^2\mathbb{I}$ , as   $\sigma_x^2, \sigma_y^2 \rightarrow 0$. CCA assumes that the emissions from the latent spaces to the observables are deterministic. This is a strong and unrealistic assumption given that word embeddings are learned from noisy corpora and stochastic learning algorithms.

\section{Experiments}

In this section, we empirically demonstrate the effectiveness of our generative approach on several benchmarks, and compare it with state-of-the-art methods. We first present cross-lingual (word-translation) evaluation tasks to evaluate the quality of our multi-lingual word embeddings. As a follow-up to the word retrieval task we also run experiments on cross-lingual sentence retrieval tasks. We further demonstrate the quality of our multi-lingual word embeddings on monolingual word- and sentence-level similarity tasks from \cite{faruqui2014improving}, which we believe provides empirical evidence that the aligned embeddings preserve and even potentially enhance their monolingual quality.

\subsection{Word Translation}

\begin{table*}[t!]
\caption{Precision @1 for cross-lingual word similarity tasks. Rows labelled AdvR are copies of Adversarial - Refine rows in \cite{conneau2017word}. Results marked with a * differ from the ones shown in \cite{conneau2017word} due to pre-processing done on their part. SVD and IBFA in the semi-supervised setting use the pseudo-dictionary, while AdvR uses frequency information. CSLS is the post-processing technique proposed in \cite{conneau2017word}.}\begin{tabular}{@{}p{2.8cm}p{.85cm}p{.85cm}|p{.8cm}p{.8cm}|p{.9cm}p{.9cm}|p{.85cm}p{.85cm}|p{.95cm}p{.95cm}@{}}
\toprule
Method  &  en-es & es-en & en-fr & fr-en & en-de & de-en & en-ru & ru-en & en-zh & zh-en \\ \midrule
\textit{Supervised}\\
\midrule
SVD &  77.4 & 77.3 & 74.9 & 76.1 & 68.4 & 67.7 & \textbf{47.0} & 58.2 & 27.3*
& 09.3*
\\ 
IBFA & \textbf{79.5} & \textbf{81.5} & \textbf{77.3} & \textbf{79.5} & \textbf{70.7} & \textbf{72.1} & 46.7 & \textbf{61.3} & \textbf{42.9} & \textbf{36.9}  \\
\midrule
SVD+CSLS & 81.4 & 82.9 & 81.1 & 82.4 & 73.5 & 72.4 & \textbf{51.7} & 63.7 & 
32.5*
& 25.1*
\\
IBFA+CSLS & \textbf{81.7} & \textbf{84.1} & \textbf{81.9} & \textbf{83.4} & \textbf{74.1} & \textbf{75.7} & 50.5 & \textbf{66.3} & \textbf{48.4} & \textbf{41.7}  \\
\midrule
\textit{Semi-supervised}\\
\midrule
SVD &  65.9 & 74.1 & 71.0 & 72.7 & 60.3 & 65.3 & 11.4 & 37.7 & 06.8 & 00.8  \\ 
IBFA & 76.1 & \textbf{80.1} & 77.1 & \textbf{78.9} & 66.8 & \textbf{71.8} & 23.1 & 39.9 & 17.1 & \textbf{24.0}  \\ 
AdvR & \textbf{79.1} & 78.1 & \textbf{78.1} & 78.2 & \textbf{71.3} & 69.6 & \textbf{37.3} & \textbf{54.3} & \textbf{30.9} & 21.9 \\
\midrule
SVD+CSLS &  73.0 & 80.7 & 75.7 & 79.6 & 65.3 & 70.8 & 20.9 & 41.5 & 10.5 & 01.7  \\ 
IBFA+CSLS & 76.5 & \textbf{83.7} & 78.6 & \textbf{82.3} & 68.7 & \textbf{73.7} & 25.3 & 46.3 & 22.1 & 27.2  \\ 
AdvR+CSLS & \textbf{81.7} & 83.3 & \textbf{82.3} & 82.1 & \textbf{74.0} & 72.2 & \textbf{44.0} & \textbf{59.1} & \textbf{32.5} & \textbf{31.4} \\
\bottomrule
\end{tabular} \label{tab:Muse}
\end{table*}

This task is concerned with the problem of retrieving the translation of a given set of source words. We reproduce results in the same environment as \cite{conneau2017word}\footnote{\href{https://github.com/Babylonpartners/MultilingualFactorAnalysis}{github.com/Babylonpartners/MultilingualFactorAnalysis}, based on \href{https://github.com/facebookresearch/MUSE}{github.com/facebookresearch/MUSE}.} for a fair comparison. We perform an ablation study to assess the effectiveness of our method in the Italian to English (it-en) setting in \cite{smith2017offline, dinu2014improving}. In these experiments we are interested in studying the effectiveness of our method compared to that of the Procrustes-based fitting used in \cite{smith2017offline} without any post-processing steps to address the hubness problem \citep{dinu2014improving}. In Table \ref{tab:Muse} we observe how our model is competitive to the results in \cite{conneau2017word} and outperforms them in most cases. We notice that given an expert dictionary, our method performs the best out of all compared methods on all tasks, except in English to Russian (en-ru) translation where it remains competitive. What is surprising is that, in the semi-supervised setting, IBFA bridges the gap between the method proposed in \cite{conneau2017word} on languages where the dictionary of identical tokens across languages (i.e. the pseudo-dictionary from \cite{smith2017offline}) is richer. However, even though it significantly outperforms SVD using the pseudo-dictionary, it cannot match the performance of the adversarial approach for more distant languages like English and Chinese (en-zh).

\subsubsection{Detailed Comparison to Basic SVD}

We present a more detailed comparison to the SVD method described in \cite{smith2017offline}. We focus on methods in their base form, that is without post-processing techniques, i.e. cross-domain similarity local scaling (CSLS) \cite{conneau2017word} or inverted softmax (ISF) \cite{smith2017offline}. Note that \cite{smith2017offline} used the scikit-learn \footnote{A commonly used Python library for scientific computing, found at \cite{scikit-learn}.} implementation of CCA, which uses an iterative estimation of partial least squares. This does not give the same results as the standard CCA procedure. In Table \ref{ablation} we reproduce the results from \cite{smith2017offline} using the dictionaries and embeddings provided by \cite{dinu2014improving}\footnote{\href{https://zenodo.org/record/2654864}{https://zenodo.org/record/2654864}} and we compare our method (IBFA) using both the expert dictionaries from \cite{dinu2014improving} and the pseudo-dictionaries as constructed in \cite{smith2017offline}. We significantly outperform both SVD and CCA, especially when using the pseudo-dictionaries.

\subsection{Word Similarity Tasks}

\begin{table*}[t!]
\captionsetup{width=\linewidth}
\caption{Comparisons without post-processing of methods. Results reproduced from \cite{smith2017offline} for fair comparison. \textbf{Left}: Comparisons using the same expert dictionary as \cite{smith2017offline}. \textbf{Right}: Comparisons using the pseudo-dictionary from \cite{smith2017offline}. \label{ablation}}
{\begin{tabular}{@{}lllllll@{}@{}lllllll@{}}
\toprule
\multirow{2}{*}{}  & \multicolumn{3}{l}{English to Italian}  & \multicolumn{3}{l}{Italian to English} & \multicolumn{3}{?l}{English to Italian}  & \multicolumn{3}{l}{Italian to English} \\
                                     & @1   & @5   & @10                       & @1          & @5          & \multicolumn{1}{l}{@10}
                                     & \multicolumn{1}{?l}{@1}   & @5   & @10                       & @1          & @5          & @10 \\ \midrule
\multicolumn{1}{l|}{Mikolov et. al.} & 33.8 & 48.3 & \multicolumn{1}{l|}{53.9} & 24.9        & 41.0        & \multicolumn{1}{l}{47.4} & \multicolumn{1}{?l}{1.0} & 2.8 & \multicolumn{1}{l|}{3.9} & 2.5        & 6.4        & 9.1     \\
\multicolumn{1}{l|}{CCA (Sklearn)}   & 36.1 & 52.7 & \multicolumn{1}{l|}{58.1} & 31.0        & 49.9        & 57.0  & \multicolumn{1}{?l}{29.1} & 46.4 & \multicolumn{1}{l|}{53.0} & 27.0        & 47.0        & 52.3    \\
\multicolumn{1}{l|}{CCA}             & 30.9 & 48.1 & \multicolumn{1}{l|}{52.7} & 27.7        & 45.5        & 51.0   & \multicolumn{1}{?l}{26.5} & 42.5 & \multicolumn{1}{l|}{48.1} & 22.8        & 40.1        & 45.5     \\
\multicolumn{1}{l|}{SVD}             &  36.9 & 52.7 & \multicolumn{1}{l|}{57.9} & 32.2       & 49.6       & 55.7   & \multicolumn{1}{?l}{27.1} & 43.4 & \multicolumn{1}{l|}{49.3} & 26.2        & 42.1        & 49.0    \\
\multicolumn{1}{l|}{IBFA (Ours)}            & \textbf{39.3} & \textbf{55.3} & \multicolumn{1}{l|}{\textbf{60.1}} & \textbf{34.7}        & \textbf{53.5}        & \textbf{59.4}    & \multicolumn{1}{?l}{\textbf{34.7}} & \textbf{52.6} & \multicolumn{1}{l|}{\textbf{58.3}} & \textbf{33.7}        & \textbf{53.3}        & \textbf{59.2}   \\ \bottomrule
\end{tabular} }
\end{table*}

\begin{table*}[t!]
\caption{Spearman correlation for English word similarity tasks. First row represents monolingual fasttext vectors \cite{JoulinGBM16fasttext} in English, the rest are bilingual embeddings. \label{tab:word-sim}}
\begin{tabular}{@{}lllllllll@{}}
\toprule
Embeddings &  WS & WS-SIM & WS-REL & RG-65 & MC-30 & MT-287 & MT-771 & MEN-TR \\ \midrule
English   & 73.7   & 78.1   & 68.2 & 79.7  & 81.2 & 67.9      & 66.9      & 76.4      \\
IBFA en-de &  \textbf{74.4}   & \textbf{79.4}   & \textbf{68.3}   & \textbf{81.4}  & \textbf{84.2}  & 67.2 & 69.4 & \textbf{77.8}     \\ 
IBFA en-fr & 72.4 & 77.8 & 65.8 & 80.5 & 83.0  & \textbf{68.2} & \textbf{69.6} & 77.6  \\ 
IBFA en-es & 73.6 & 78.5 & 67.0 & 79.0 & 83.0  & 68.2 & 69.4  & 77.3 \\ 
CCA en-de &  71.7   & 76.4  & 64.0 & 76.7 & 82.4 & 63.0 & 64.7 & 75.3 \\ 
CCA en-fr & 70.9   & 76.4 & 63.3 & 76.5 & 81.4 & 63.4 & 65.4 & 74.9 \\ 
CCA en-es & 70.8   & 76.3   & 63.1   & 76.4 & 81.2 & 63.0      & 65.1 & 74.7    \\ 
\bottomrule
\end{tabular}
\end{table*}

This task assesses the monolingual quality of word embeddings. In this experiment, we fit both considered methods (CCA and IBFA) on the entire available dictionary of around 100k word pairs. We compare to CCA as used in \cite{faruqui2014improving} and standard monolingual word embeddings on the available tasks from \cite{faruqui2014improving}. We evaluate our multilingual embeddings on the following tasks: \textbf{WS353} \cite{finkelstein2002placing}; \textbf{WS-SIM}, \textbf{WS-REL} \cite{agirre2009study}; \textbf{RG65} \cite{rubenstein1965contextual};
\textbf{MC-30} \cite{miller1991contextual}; \textbf{MT-287}; \cite{radinsky2011word}; \textbf{MT-771} \cite{halawi2012large}, and \textbf{MEN-TR} \cite{bruni2012distributional}. These tasks consist of English word pairs that have been assigned ground truth similarity scores by humans. We use the test-suite provided by \cite{faruqui-2014:SystemDemo}\footnote{\href{https://github.com/mfaruqui/eval-word-vectors}{https://github.com/mfaruqui/eval-word-vectors}} to evaluate our multilingual embeddings on these datasets. This test-suite calculates similarity of words through cosine similarity in their representation spaces and then reports Spearman correlation with the ground truth similarity scores provided by humans.

As shown in Table \ref{tab:word-sim}, we observe a performance gain over CCA and monolingual word embeddings suggesting that we not only preserve the monolingual quality of the embeddings but also enhance it. 

\subsection{Monolingual Sentence Similarity Tasks}

Semantic Textual Similarity (STS) is a standard benchmark used to assess sentence similarity metrics \citep{agirre2012semeval, agirre2013sem, agirre2014semeval, agirre2015semeval, agirre2016semeval}. In this work, we use it to show that our alignment procedure does not degrade the quality of the embeddings at the sentence level. For both IBFA and CCA, we align English and one other language (from French, Spanish, German) using the entire dictionaries (of about 100k word pairs each) provided by \cite{conneau2017word}. We then use the procedure defined in \cite{arora2016simple} to create sentence embeddings and use cosine similarity to output sentence similarity using those embeddings. The method's performance on each set of embeddings is assessed using Spearman correlation to human-produced expert similarity scores. As evidenced by the results shown in Table \ref{tab:sts}, IBFA remains competitive using any of the three languages considered, while CCA shows a performance decrease.

\subsection{Crosslingual Sentence Similarity Tasks}

\begin{table*}[t!]
\caption{Spearman correlation for Semantic Textual Similarity (STS) tasks in English. All results use the sentence embeddings described in \cite{arora2016simple}. First row represents monolingual FastText vectors \cite{JoulinGBM16fasttext} in English, the rest are bilingual embeddings. *STS13 excludes the proprietary SMT dataset. \label{tab:sts}}
\begin{tabular}{@{}llllll@{}}
\toprule
Embeddings &  STS12 & STS13* & STS14 & STS15 & STS16 \\ \midrule
English    & \textbf{58.1} & 69.2 & 66.7 & 72.6 & 70.6 \\
IBFA en-de & \textbf{58.1} & \textbf{70.2} & \textbf{66.8} & 73.0 & 71.6 \\
IBFA en-fr & 58.0 & 70.0 & 66.7 & 72.8 & 71.4 \\ 
IBFA en-es & 57.9 & 69.7 & 66.6 & 72.9 & \textbf{71.7} \\ 
CCA en-de  & 56.7 & 67.5 & 65.7 & \textbf{73.1} & 70.5 \\ 
CCA en-fr  & 56.7 & 67.9 & 65.9 & 72.8 & 70.8 \\ 
CCA en-es  & 56.6 & 67.8 & 65.9 & 72.9 & 70.8 \\ 
\bottomrule
\end{tabular}
\end{table*}

\begin{table*}[t!]
\captionsetup{width=\linewidth}
\caption{Sentence translation precisions @1, @5, @10 on 2,000 English-Italian pairs samples from a set of 200k sentences from Europarl \cite{koehn2005europarl} on Dinu embeddings. AdvR is copied from Adversarial - Refined in \cite{conneau2017word}. Rows with $\checkmark$ copied from \cite{smith2017offline}. } \label{tab:sentence-precision}
{\begin{tabular}{@{}lllllll@{}}
\toprule
\multirow{2}{*}{}                    & \multicolumn{3}{l}{English to Italian}  & \multicolumn{3}{l}{Italian to English} \\
                                     & @1   & @5   & @10                       & @1          & @5          & @10        \\ \midrule
\multicolumn{1}{l|}{Mikolov et. al.$\checkmark$} & 10.5 & 18.7 & \multicolumn{1}{l|}{22.8} & 12.0        & 22.1        & 26.7       \\
\multicolumn{1}{l|}{Dinu et al.$\checkmark$}   & 45.3 & 72.4 & \multicolumn{1}{l|}{80.7} & 48.9        & 71.3        & 78.3       \\
\multicolumn{1}{l|}{Smith et al.$\checkmark$}             & 54.6 & 72.7 & \multicolumn{1}{l|}{78.2} & 42.9        & 62.2        & 69.2       \\
\multicolumn{1}{l|}{SVD} & 40.5 & 52.6 & \multicolumn{1}{l|}{56.9} & 51.2 & 63.7 & 67.9 \\
\multicolumn{1}{l|}{IBFA (Ours)} & 62.7 & 74.2 & \multicolumn{1}{l|}{77.9} & 64.1 &
75.2 & 79.5 \\
\multicolumn{1}{l|}{SVD + CSLS} & 64.0 & 75.8 & \multicolumn{1}{l|}{78.5} & 67.9 & 79.4 & 82.8 \\
\multicolumn{1}{l|}{AdvR + CSLS} & 66.2 & 80.4 & \multicolumn{1}{l|}{83.4} & 58.7 & 76.5 & 80.9
\\
\multicolumn{1}{l|}{IBFA + CSLS} & \textbf{68.8} & \textbf{80.7} & \multicolumn{1}{l|}{\textbf{83.5}} & \textbf{70.2} &
\textbf{80.8} & \textbf{84.8} \\
 \bottomrule
\end{tabular} }
\end{table*}

Europarl \citep{koehn2005europarl} is a parallel corpus of sentences taken from the proceedings of the European parliament. In this set of experiments, we focus on its English-Italian (en-it) sub-corpus, in order to compare to previous methods. We report results under the framework of \cite{conneau2017word}. That is, we form sentence embeddings using the average of the tf-idf weighted word embeddings in the bag-of-words representation of the sentence. Performance is averaged over 2,000 randomly chosen source sentence queries and 200k target sentences for each language pair. Note that this is a different set up to the one presented in \cite{smith2017offline}, in which an unweighted average is used. The results are reported in Table \ref{tab:sentence-precision}. As we can see, IBFA outperforms all prior methods both using nearest neighbour retrieval, where it has a gain of 20 percent absolute on SVD, as well as using the CSLS retrieval metric.

\subsection{Alignment of three languages}

\begin{table*}[t!]
\caption{Precision @1 when aligning English, French and Italian embeddings to a common space. For SVD, this common space is English, while for MBFA it is the latent space. \label{tab:3-language-space}}
\begin{tabular}{@{}lllllll@{}}
\toprule
Method    & en-it & it-en & en-fr & fr-en & it-fr & fr-it \\ 
\midrule
SVD       &  71.0 & 72.4 & 74.9 & 76.1 & 78.3 & 72.9      \\
MBFA      &  \textbf{71.9} & \textbf{73.4} & \textbf{76.7} & \textbf{78.1} & \textbf{82.6} & \textbf{77.5}     \\ 
\midrule
SVD+CSLS  &  76.2 & \textbf{77.9} & 81.1 & \textbf{82.4} & 84.5 & 79.8 \\ 
MBFA+CSLS & \textbf{77.4} & 77.7 & \textbf{81.9} & 82.1 & \textbf{86.8} & \textbf{81.9} \\
\bottomrule
\end{tabular}
\end{table*}

In an ideal scenario, when we have $v$ languages, we wouldn't want to train a transformation between each pair, as that would involve storing $\mathcal{O}(v^2)$ matrices. One way to overcome this problem is by aligning all embeddings to a common space. In this exploratory experiment, we constrain ourselves to aligning three languages at the same time, but the same methodology could be applied to an arbitrary number of languages. MBFA, the extension of IBFA described in Section \ref{subsec:mbfa} naturally lends itself to this task. What is needed for training this method is a dictionary of word triples across the three languages considered. We construct such a dictionary by taking the intersection of all 6 pairs of bilingual dictionaries for the three languages provided by \cite{conneau2017word}. We then train MBFA for 20,000 iterations of EM (a brief analysis of convergence is provided in Appendix \ref{apdx:curve}). Alternatively, with direct methods like \cite{smith2017offline, conneau2017word} one could align all languages to English and treat that as the common space.

We compare both approaches and present their results in Table \ref{tab:3-language-space}. As we can see, both methods experience a decrease in overall performance when compared to models fitted on just a pair of languages, however MBFA performs better overall. That is, the direct approaches preserve their performance on translation to and from English, but translation from French to Italian decreases significantly. Meanwhile, MBFA suffers a decrease in each pair of languages, however it retains competitive performance to the direct methods on English translation. It is worth noting that as the number of aligned languages $v$ increases, there are $O(v)$ pairs of languages, one of which is English, and $O(v^2)$ pairs in which English does not participate. This suggests that MBFA may generalise past three simultaneously aligned languages better than the direct methods.

\subsection{Generating Random Word Pairs}

We explore the generative process of IBFA by synthesising word pairs from noise, using a trained English-Spanish IBFA model. We follow the generative process specified in Equation \ref{eq:generative} to generate 2,000 word vector pairs and then we find the nearest neighbour vector in each vocabulary and display the corresponding words. We then rank these 2,000 pairs according to their joint probability under the model and present the top 28 samples in Table \ref{tab:samps}. Note that whilst the sampled pairs are not exact translations, they have closely related meanings. The examples we found interesting are dreadful and despair; frightening and brutality; crazed and merry; unrealistic and questioning; misguided and conceal; reactionary and conservatism.

\renewcommand{\arraystretch}{1.14}
\begin{table}
    \caption{Random pairs sampled from model, selected top 28 ranked by confidence. Proper nouns, and acronyms (names and surnames) were removed from the list. Third column represents a correct translation from Spanish to English.}
    \begin{tabular}{@{}p{2.2cm}p{2.6cm}p{2.1cm}@{}}
    \toprule
    en & es & es$\rightarrow$en\\
    \midrule
         particular & efectivamente & effectively \\
         correspondingly & esto & this\\
         silly & irónicamente & ironic \\
         frightening & brutalidad & brutality \\
         manipulations & intencionadamente & intentionally \\
         ignore & contraproducente & counter-productive\\
         fundamentally & entendido & understood\\
         embarrassed & enojado & angry \\
         terrified & casualidad & coincidence \\
         hypocritical & obviamente & obviously\\
         wondered & incómodo & uncomfort-able \\
         oftentimes & apostar & betting \\
         unwittingly & traicionar & betray \\
         mishap & irónicamente & ironically  \\
         veritable & empero & however \\
         overpowered & deshacerse & fall apart\\
         crazed & divertidos & merry \\
         frightening & ironía & irony \\
         dreadful & desesperación & despair\\
         instituting & restablecimiento & recover \\
         unrealistic & cuestionamiento & questioning \\
         regrettable & erróneos & mistaken \\
         irresponsible & preocupaciones & concerns\\
         obsession & irremediablemente & hopelessly\\
         embodied & voluntad  & will\\
         misguided & esconder & conceal \\
         perspective & contestación & answer \\
         reactionary & conservadurismo & conservatism \\
    \bottomrule
    \end{tabular}
    \label{tab:samps}
\end{table}

\section{Conclusion}
We have introduced a cross-lingual embedding alignment procedure based on a probabilistic latent variable model, that increases performance across various tasks compared to previous methods using both nearest neighbour retrieval, as well as the CSLS criterion. We have shown that the resulting embeddings in this aligned space preserve their quality by presenting results on tasks that assess word and sentence-level monolingual similarity correlation with human scores. The resulting embeddings also significantly increase the precision of sentence retrieval in multilingual settings. 
Finally, the preliminary results we have shown on aligning more than two languages at the same time provide an exciting path for future research.

\bibliography{acl2019}
\bibliographystyle{acl_natbib}

\appendix
\section{Joint Distribution}

We show the form of the joint distribution for 2 views. Concatenating our data and parameters as below, we can use Equation (3) of \cite{ghahramani1996algorithm} to write 
\begin{align}
    \bm{m} &= \left[ \begin{array} { l }
        \bm {x} \\
        \bm {y}
    \end{array} \right], \bm{W} = \left[ \begin{array} { l }
        \bm {W}_x \\
        \bm {W}_y
    \end{array} \right] \nonumber \\
    \bm{\Psi} &= \left[
    \begin{array} { c c } 
         \bm { \bm{\Psi}_x } & \bm{0} \nonumber \\
         \bm{0} & \bm { \bm{\Psi}_y }
    \end{array} \right], \bm{\mu} = \left[
    \begin{array} { l } 
         \bm { \bm{\mu}_x }  \\
         \bm { \bm{\mu}_y }
    \end{array} \right] \\
    p(\bm{m}, \bm{z} | \bm{\theta}) &= \mathcal { N } \left( 
    \left[
    \begin{array} { l }
        \bm { m } \\
        \bm { z }
    \end{array} \right] ; 
       \left[
    \begin{array} { l } 
         \bm { \mu }  \\
         \bm { 0 }
    \end{array} \right] , \bm{\Sigma}_{m, z} \right) \label{joint-gaussian} \\
    \bm{\Sigma}_{m, z} &= \left[
    \begin{array} { c c } 
         \bm { \bm{W} \bm{W}^{\top} + \bm{\Psi} } & \bm{W} \nonumber \\
         \bm{W}^\top & \bm { \mathbb{I} }
    \end{array} \right]
\end{align}
It is clear that this generalises to any number of views of any dimension, as the concatenation operation does not make any assumptions.

\section{Projections to Latent Space $ \mathbb{E}_{p(\vz|\vx)}[\vz]$}
\label{apdx:marginal}
We can query the joint Gaussian in \ref{joint-gaussian} using rules from \cite{petersen2008matrix} Sections (8.1.2, 8.1.3) and we get
\begin{align}
    p(\bm{z} | \bm{x}) &= \mathcal{N} \left(\bm{z}; \bm{W}_x^{\top} \bm{\Sigma}_{x}^{-1} \tilde{\bm{x}}, \mathbb{I} - \bm{W}_x^{\top} \bm{\Sigma}_{x}^{-1} \bm{W}_x \right) \nonumber \\
    \mathbb{E}[\bm{z}|\bm{x}] &= \bm{W}_x^{\top} \bm{\Sigma}_{x}^{-1} \tilde{\bm{x}} \nonumber
\end{align}

\section{Derivation for the Marginal Likelihood}
We want to compute $p(\bm{x}, \bm{y} | \bm{\theta})$ so that we can then learn the parameters $\bm{\theta} = \{\bm{\theta}_x, \bm{\theta}_y\}$, $\bm{\theta}_i = \{\bm{\mu}_i, \bm{W}_{i}, \bm{\Psi}_{i},  \}$ by maximising the marginal likelihood as is done in Factor Analysis. 

From the joint $p(\bm{m}, \bm{z} | \bm{\theta})$, again using rules from \cite{petersen2008matrix} Sections (8.1.2) we get
\begin{align}
    p(\bm{m} | \bm{\theta}) &= p(\bm{x}, \bm{y} | \bm{\theta}) \nonumber \\
    &= \mathcal { N } \left( 
    \left[
    \begin{array} { l }
        \bm { x } \\
        \bm { y }
    \end{array} \right] ; 
       \left[
    \begin{array} { l } 
         \bm { \mu }_{x}  \\
         \bm { \mu }_{y}
    \end{array} \right] , \bm{W}\bm{W}^T + \bm{\Psi} \right) \nonumber
\end{align}
For the case of two views, the joint probability can be factored as
\begin{align*}
    p(\bm{x}, \bm{y} | \bm{\theta}) &= p(\bm{x}| \bm{\theta}_x) p(\bm{y} | \bm{x}, \bm{\theta}) \nonumber \\
    p(\bm{x}| \bm{\theta}_x) &= \mathcal { N } \left(\bm{x}; \vmu_x, \bm{\Sigma}_x \right) \nonumber \\
    p(\bm{y} | \bm{x}, \bm{\theta}) &= \mathcal { N } \left(\bm{y}; \bm{W}_y \bm{W}_x^{\top}
    \bm{\Sigma}_{x}^{-1} \tilde{\bm{x}} + \bm{\mu}_{y}, \bm{\Sigma}_{\bm{y}|\bm{x}} \right) \\
    &= \mathcal { N } \left(\bm{y}; \bm{W}_y E[\bm{z}|\bm{x}] + \bm{\mu}_{y}, \bm{\Sigma}_{\bm{y}|\bm{x}} \right), \\
\end{align*}
where
\begin{align*}
    \bm{\Sigma}_{x} &= \bm{W}_x\bm{W}_x^\top + \bm{\Psi}_x \\
    \bm{\Sigma}_{\bm{y}|\bm{x}} &= \bm{\Sigma}_{y} - \bm{W}_{y}\bm{W}_{x}^{\top}\bm{\Sigma}_x^{-1}\bm{W}_{x} \bm{W}_{y}^{\top}
\end{align*}
\section{Scaled Reconstruction Errors}

\label{apdx:reconstruction}
\begin{align}
    \log p(\bm{x}, \bm{y} | \bm{\theta}) &= \log p^*(\bm{x}| \bm{\theta}_x) + \log p^*(\bm{y} | \bm{x}, \bm{\theta}) \nonumber \\
    & - \frac{1}{2}(\log |2 \pi \mSigma_{\vy|\vx}| + \log |2 \pi \mSigma_{\vx}|) \nonumber \\
    \log p^*(\bm{y} |  \bm{x}, \bm{\theta}) &= -\frac{1}{2} ||\tilde{\bm{y}} - \bm{W}_y E[\bm{z}|\bm{x}] ||_{\bm{\Sigma}_{y | x}}^2 \nonumber \noindent \\
    \log p^*(\bm{x}| \bm{\theta}_x) &= -\frac{1}{2} ||\bm{x} - \bm{\mu}_x||_{\bm{\Sigma}_{x}}^2 \nonumber \\
    &= -\frac{1}{2} ||\bm{\Sigma}_{x}^{-\frac{1}{2}}\tilde{\bm{x}}||^2 \nonumber
\end{align}
Setting $\bm{A} = \bm{\Psi}_x \bm{\Sigma}_x^{-1} \bm{\Psi}_x$, we can re-parametrise as
\begin{align}
    \log p^*(\bm{x}| \bm{\theta}_x) &= -\frac{1}{2} ||\bm{\Psi}_x \bm{\Sigma}_{x}^{-1}\tilde{\bm{x}}||_{\bm{A}}^2 \nonumber \\
    &= -\frac{1}{2} ||(\bm{\Sigma}_{x} - \bm{W}_x\bm{W}_x^\top) \bm{\Sigma}_{x}^{-1}\tilde{\bm{x}}||_{\bm{A}}^2 \nonumber \\
    &= -\frac{1}{2} ||\tilde{\bm{x}} - \bm{W}_x\bm{W}_x^\top \bm{\Sigma}_{x}^{-1}\tilde{\bm{x}}||_{\bm{A}}^2 \nonumber \\
    &= -\frac{1}{2} ||\tilde{\bm{x}} - \bm{W}_xE[\bm{z}|\bm{x}]||_{\bm{A}}^2 \nonumber
\end{align}
\section{Expectation Maximisation for MBFA}
\label{apdx:em}
Define
\begin{align*}
    \tilde{\bm{x}} = \begin{bmatrix}
                        \bm{x}_1 - \vmu_1\\ 
                        \vdots\\ 
                        \bm{x}_v - \vmu_1
                    \end{bmatrix},
    \mW = \begin{bmatrix}
                        \mW_1\\ 
                        \vdots\\ 
                        \mW_v
                    \end{bmatrix}
\end{align*}
\begin{align*}
    \bm{\Psi} = \begin{bmatrix}
                \bm{\Psi}_1 &  & 0\\ 
                 & \ddots & \\ 
                 0 &  & \bm{\Psi}_v
            \end{bmatrix} = \text{Bdiag}(\bm{\Psi}_1, \hdots, \bm{\Psi}_v)
\end{align*}
Hence
\begin{align*}
    p(\tilde{\bm{x}} | \bm{z}; \bm{\Psi}, \bm{W}) =  \mathcal{N}(\tilde{\bm{x}} | \bm{W}\bm{z}, \bm{\Psi})
\end{align*}

\begin{table*}[t!]
\centering
\caption{Precision @1 between MBFA fitted for 1K iterations and MBFA fitted for 20K iterations. \label{tab:em-precision}}
\begin{tabular}{@{}lllllll@{}}
\toprule
Method    & EN-IT & IT-EN & EN-FR & FR-EN & IT-FR & FR-IT \\ 
\midrule
MBFA-1K       &  71.9 & 73.3 & 76.7 & 78.2 & 82.4 & 77.5 \\ 
MBFA-20K      &  71.9 & 73.4 & 76.7 & 78.1 & 82.6 & 77.5 \\ 
\midrule
MBFA-1K+CSLS  & 77.5 & 77.6 & 81.9 & 82.0 & 86.8 & 82.1  \\
MBFA-20K+CSLS & 77.4 & 77.7 & 81.9 & 82.1 & 86.8 & 81.9  \\
\bottomrule
\end{tabular}
\end{table*}

\noindent This follows the same form as regular factor analysis, but with a block-diagonal constraint on $\bm{\Psi}$. Thus by Equations (5) and (6) of \cite{ghahramani1996algorithm}, we apply EM as follows.\\

\noindent\textbf{E-Step:} Compute $\mathbb{E}[ \bm{z} | \bm{x} ]$ and $\mathbb{E}[ \bm{zz^\top} | \bm{x} ]$ given the parameters $\bm{\theta}_t = \{ \bm{W}_t, \bm{\Psi}_t \}$.
\begin{align}
    \mathbb{E}[ \bm{z}^{(i)} | \tilde{\bm{x}}^{(i)} ] &= \bm{B}_t \tilde{\bm{x}}^{(i)} \nonumber \\
    \mathbb{E}[ \bm{z}^{(i)}\bm{z}^{(i)^\top} | \tilde{\bm{x}}^{(i)} ] &= \mathbb{I} - \bm{B}_t \bm{W}_t + \bm{B}_t \tilde{\bm{x}}^{(i)} \tilde{\bm{x}}^{(i)\top} \bm{B}_t^\top \nonumber \\
    &= \bm{M}_t + \bm{B}_t\tilde{\bm{x}}^{(i)} \tilde{\bm{x}}^{(i)\top}\bm{B}_t^\top \label{eq:woodbury}
\end{align}
\noindent where
\begin{align}
    \bm{M}_t &= \left( \mathbb{I} + \bm{W}_t^\top \bm{\Psi}_t^{-1}\bm{W}_t\right)^{-1} \nonumber \\
    \bm{B}_t &= \bm{W}_t^\top (\bm{\Psi}_t + \bm{W}_t\bm{W}_t^\top)^{-1} \nonumber \\
    &= \bm{M}_t\bm{W}_t^{\top}\bm{\Psi}_t^{-1}. \label{eq:push-through}
\end{align}

\noindent Equation \ref{eq:woodbury} is obtained by applying the Woodbury identity, and Equation \ref{eq:push-through} by applying the closely related push-through identity, as found in Section 3.2 of \cite{petersen2008matrix}.\\

\noindent\textbf{M-Step:} Update parameters $\bm{\theta}_{t\!+\!1}\!=\!\{ \bm{W}_{t\!+\!1}, \bm{\Psi}_{t\!+\!1} \}$.\\

\noindent Define
\begin{align*}
    \mS &= \frac{1}{m}\sum\limits_{i=1}^m \tilde{\bm{x}}^{(i)} \tilde{\bm{x}}^{(i)\top}
\end{align*}
By first observing
\begin{align*}
     \frac{1}{m}\sum\limits_{i=1}^m\!  &  \tilde{\bm{x}}^{(i)} \mathbb{E}[ \bm{z}^{(i)} | \tilde{\bm{x}}^{(i)} ]^\top\! = \mS\bm{B}_t^\top \\
     \frac{1}{m}\sum\limits_{j=1}^m\!  &  \mathbb{E}[ \bm{z}^{(j)}\bm{z}^{(j)^\top} | \tilde{\bm{x}}^{(j)} ] = \bm{M}_t + \bm{B}_t\mS\bm{B}_t^\top,
\end{align*}
update the parameters as follows.
\begin{align*}
    \bm{W}_{t\!+\!1}\!\! &=\mS\bm{B}_t^\top \!\! \left( \mathbb{I} - \bm{B}_t\bm{W}_t + \bm{B}_t\mS\bm{B}_t^\top \right)^{-1} \\
    %
    %
    &=\mS\bm{B}_t^\top \!\!\left( \bm{M}_t + \bm{B}_t\mS\bm{B}_t^\top \right)^{-1} \\
    %
    %
    \widetilde{\bm{\Psi}}_{t\!+\!1}\! &=\!\!\frac{1}{m}\sum\limits_{i=1}^m \tilde{\bm{x}}^{(i)} \tilde{\bm{x}}^{(i)\top}\!\!-\!\!\bm{W}_{t\!+\!1} \mathbb{E}[ \bm{z}^{(i)} | \tilde{\bm{x}}^{(i)} ] \tilde{\bm{x}}^{(i)\top}\!\! \\
    %
    %
    &=\mS - \frac{1}{m}\sum\limits_{i=1}^m \bm{W}_{t\!+\!1} \bm{B}_t \tilde{\bm{x}}^{(i)} \tilde{\bm{x}}^{(i)\top} \\
    &=\mS - \bm{W}_{t\!+\!1} \bm{B}_t \mS \\
    &=\mS - \mS \bm{B}_t^{\top} \bm{W}_{t\!+\!1}^{\top}
\end{align*}
Imposing the block diagonal constraint,
\begin{align*}
    \bm{\Psi}_{t\!+\!1} &=\text{Bdiag}\left((\widetilde{\bm{\Psi}}_{t+1})_{11}, \dots, (\widetilde{\bm{\Psi}}_{t+1})_{vv} \right)
\end{align*}
where $(\tilde{\bm{\Psi}})_{ii} = \bm{\Psi}_{i}$.

\section{Independence to Noise in Direct Methods}
\label{apdx:ls}

We are maximising the following quantity with respect to $\bm{\theta} = \{ \bm{W}, \bm{\mu}, \bm{\Psi} \}$

\begin{align*}
    p(\bm{Y} | \bm{X}, \bm{\theta})&\!=\!\prod_i p(\bm{y}^{(i)} | \bm{x}^{(i)}, \bm{\theta}) \\
    & \!= \!\prod_i \mathcal{N} (\bm{y}^{(i)} ; \bm{W} \bm{x}^{(i)} + \bm{\mu}, \bm{\Psi}) \\
    \log p(\bm{Y} | \bm{X}, \bm{\theta})\!&=\!\!-\frac{1}{2} \!\!\left( \!\sum_i |\!|\bm{y}^{(i)} \!- \!\bm{W} \bm{x}^{(i)}|\!|_{\bm{\Psi}}^{2}\!-\!C\!\!\right) \\
\end{align*}
Then the partial derivative $\mathcal{Q} = \frac{\partial \log p(\bm{Y} | \bm{X}, \bm{\theta})}{\partial \bm{W}}$ is proportional to
\begin{align*}
    \mathcal{Q} &\propto \!\!\left ( \!\sum_i \!\bm{\Psi}^{-1}(\bm{y}^{(i)}\!-\!\bm{W} \bm{x}^{(i)}) \bm{x}^{(i)\top} \right) \\
    &\propto \!\bm{\Psi}^{-1}\!\!\left(\!\!\sum_i \!\bm{y}^{(i)}\bm{x}^{(i)\top}\!\!\!\!-\!\bm{W}\!\sum_i\!\!\bm{x}^{(i)} \bm{x}^{(i)\top}\!\!\right)
\end{align*}

\begin{figure*}[t!]
    \centering
    \includegraphics[width=.4\textwidth]{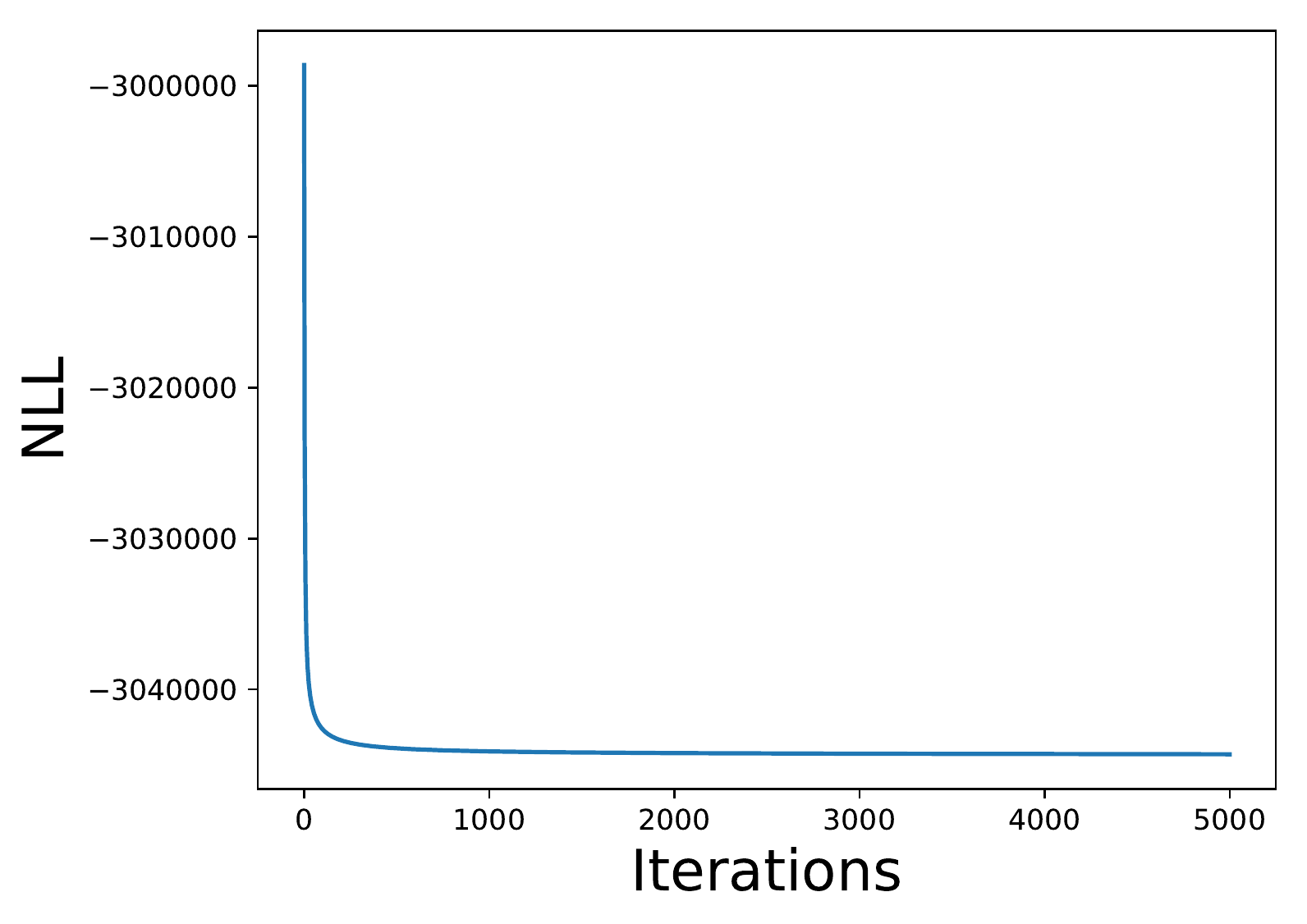}
    \includegraphics[width=.4\textwidth]{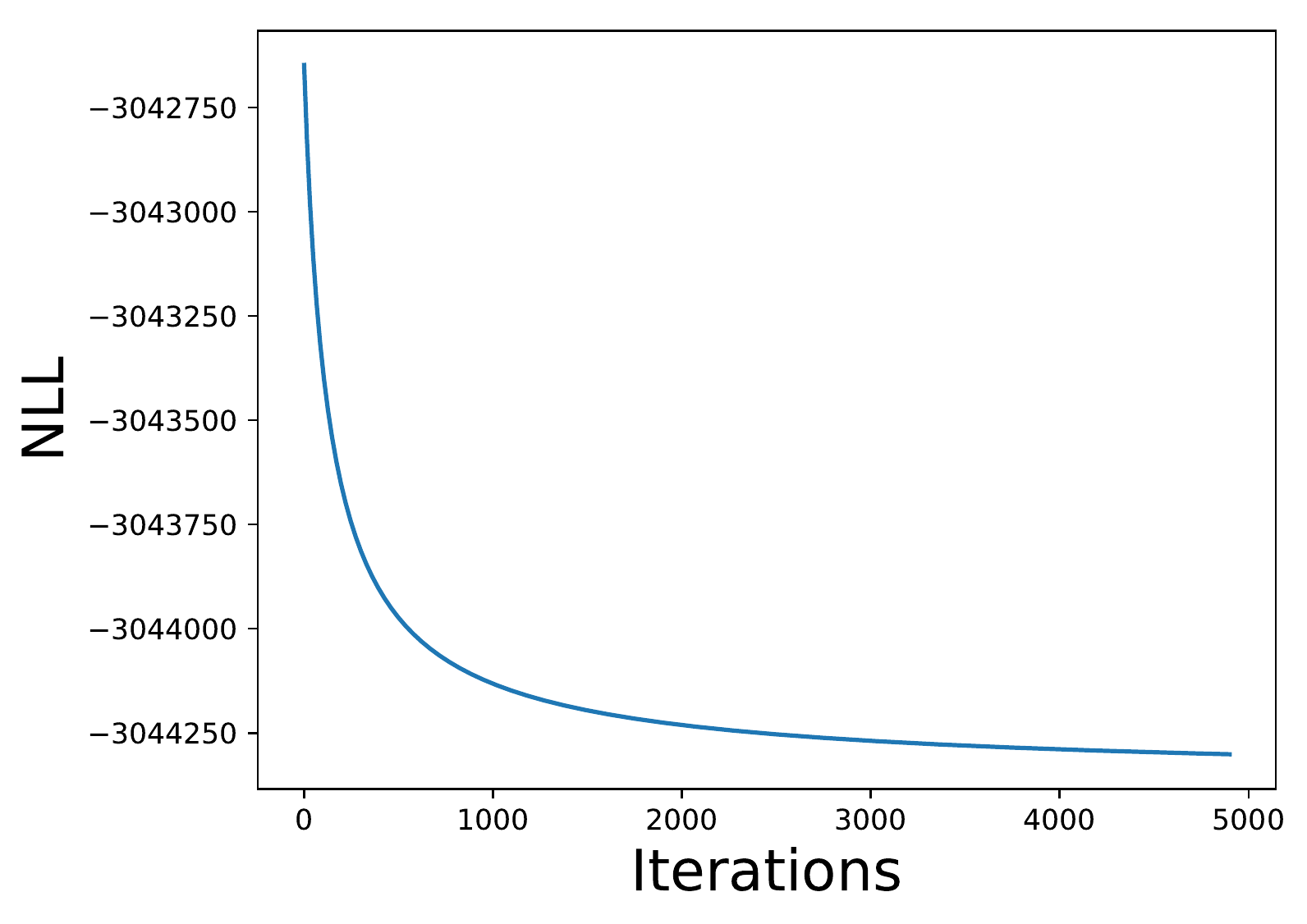}
    \caption{Training curve of EM algorithm over the first 5,000 iterations. It is clear that the procedure quickly finds a good approximation to the optimal parameters and then slowly converges to the real optimum. Left picture shows the entire training curve, while the right picture starts from iteration 100.}
    \label{fig:curve}
\end{figure*}
The maximum likelihood is achieved when
\begin{align*}
    \frac{\partial \log p(\bm{Y} | \bm{X}, \bm{\theta})}{\partial \bm{W}} &= \bm{0},
\end{align*}

\noindent and since $\bm{\Psi}^{-1}$ has an inverse (namely $\bm{\Psi}$), this means that

\begin{align*}
    \bm{W}\sum_i \bm{x}^{(i)} \bm{x}^{(i)\top}=\sum_i \bm{y}^{(i)}\bm{x}^{(i)\top}
\end{align*}

It is clear from here that the MLE of $\bm{W}$ does not depend on $\bm{\Psi}$, thus we can conclude that adding a noise parameter to this directed linear model has no effect on its predictions.

\section{Learning curve of EM} \label{apdx:curve}

Figure \ref{fig:curve} shows the negative log-likelihood of the three language model over the first 5,000 iterations. The precision of the learned model is very close when evaluated at iteration 1,000 and at iteration 20,000 as seen in Table \ref{tab:em-precision}. This suggests that the model need not be trained to full convergence to work well.

\end{document}